\documentclass[lettersize,journal]{IEEEtran}
\usepackage{amsmath,amsfonts}
\usepackage{algorithmic}
\usepackage{algorithm}
\usepackage{array}
\usepackage[caption=false,font=normalsize,labelfont=sf,textfont=sf]{subfig}
\usepackage{textcomp}
\usepackage{stfloats}
\usepackage{url}
\usepackage{verbatim}
\usepackage{graphicx}
\usepackage{cite}
\usepackage{balance}

\usepackage{cite}
\usepackage{bm} 
\usepackage{multirow} 
\usepackage{booktabs} 
\usepackage{color}
\definecolor{ggray}{RGB}{127,127,127}

\newcommand{\eg}{\textit{e}.\textit{g}.}  
\newcommand{\ie}{\textit{i}.\textit{e}.}  
\newcommand{\tabincell}[2]{\begin{tabular}{@{}#1@{}}#2\end{tabular}} 
\newcommand{\etal}{\textit{et al.}}  
\usepackage{url}  
\usepackage{colortbl}
\usepackage{makecell}
\definecolor{mygray}{gray}{.9}
\newcommand{\red}[1]{\textcolor{black}{#1}}

\newcommand{\blue}[1]{\textcolor{black}{#1}}
\newcommand{\lightgreen}[1]{\textcolor{black}{#1}}
\usepackage{bbding}
\usepackage{times}
\usepackage{epsfig}
\usepackage{multirow}

\usepackage{caption}
\usepackage{comment}
\usepackage{amsmath}
\usepackage{amssymb}
\usepackage[dvipsnames,svgnames,x11names]{xcolor}
\definecolor{citecolor}{RGB}{119,185,0} 
\usepackage[pagebackref=false,breaklinks=true,letterpaper=true,colorlinks,citecolor=citecolor,bookmarks=false]{hyperref}
\usepackage{pifont}

\def\eg{\emph{e.g.}} 
\def\ie{\emph{i.e.}} 
\def\etal{\emph{et~al.}}

\begin{document}

\title{Knowledge-guided Causal Intervention for Weakly-supervised Object Localization}

\author{Feifei Shao,
        Yawei Luo$^*$,
        Fei Gao,
        Yi Yang,
        Jun Xiao

\thanks{$^*$Yawei Luo is the corresponding author.}
\thanks{Feifei Shao, Yawei Luo, Yi Yang, Jun Xiao are with Zhejiang University, Hangzhou, China. Email: sff@zju.edu.cn, yaweiluo@zju.edu.cn, yangyics@zju.edu.cn, junx@cs.zju.edu.cn.}
\thanks{Fei Gao is with the Zhejiang University of Technology, Hangzhou, China. Email: feig@zjut.edu.cn.}
}



\maketitle

\begin{abstract}
Previous weakly-supervised object localization (WSOL) methods aim to expand activation map discriminative areas to cover the whole objects, yet neglect two inherent challenges when relying solely on image-level labels. First, the ``entangled context'' issue arises from object-context co-occurrence (\eg, fish and water), making the model inspection hard to distinguish object boundaries clearly. Second, the ``C-L dilemma'' issue results from the information decay caused by the pooling layers, which struggle to retain both the semantic information for precise classification and those essential details for accurate localization, leading to a trade-off in performance. In this paper, we propose a knowledge-guided causal intervention method, dubbed KG-CI-CAM, to address these two under-explored issues in one go. More specifically, we tackle the co-occurrence context confounder problem via causal intervention, which explores the causalities among image features, contexts, and categories to eliminate the biased object-context entanglement in the class activation maps. Based on the disentangled object feature, we introduce a multi-source knowledge guidance framework to strike a balance between absorbing classification knowledge and localization knowledge during model training. Extensive experiments conducted on several benchmark datasets demonstrate the effectiveness of KG-CI-CAM in learning distinct object boundaries amidst confounding contexts and mitigating the dilemma between classification and localization performance.\footnote{Our code is publicly available at \url{https://github.com/shaofeifei11/KG-CI-CAM}} 
\end{abstract}

\begin{IEEEkeywords}
    Object Localization, Weakly-supervised Learning, Knowledge Guidance, Causal Intervention.
\end{IEEEkeywords}

\section{Introduction}
\begin{figure*}[t]
   \centering
   \includegraphics[width=1.0\linewidth]{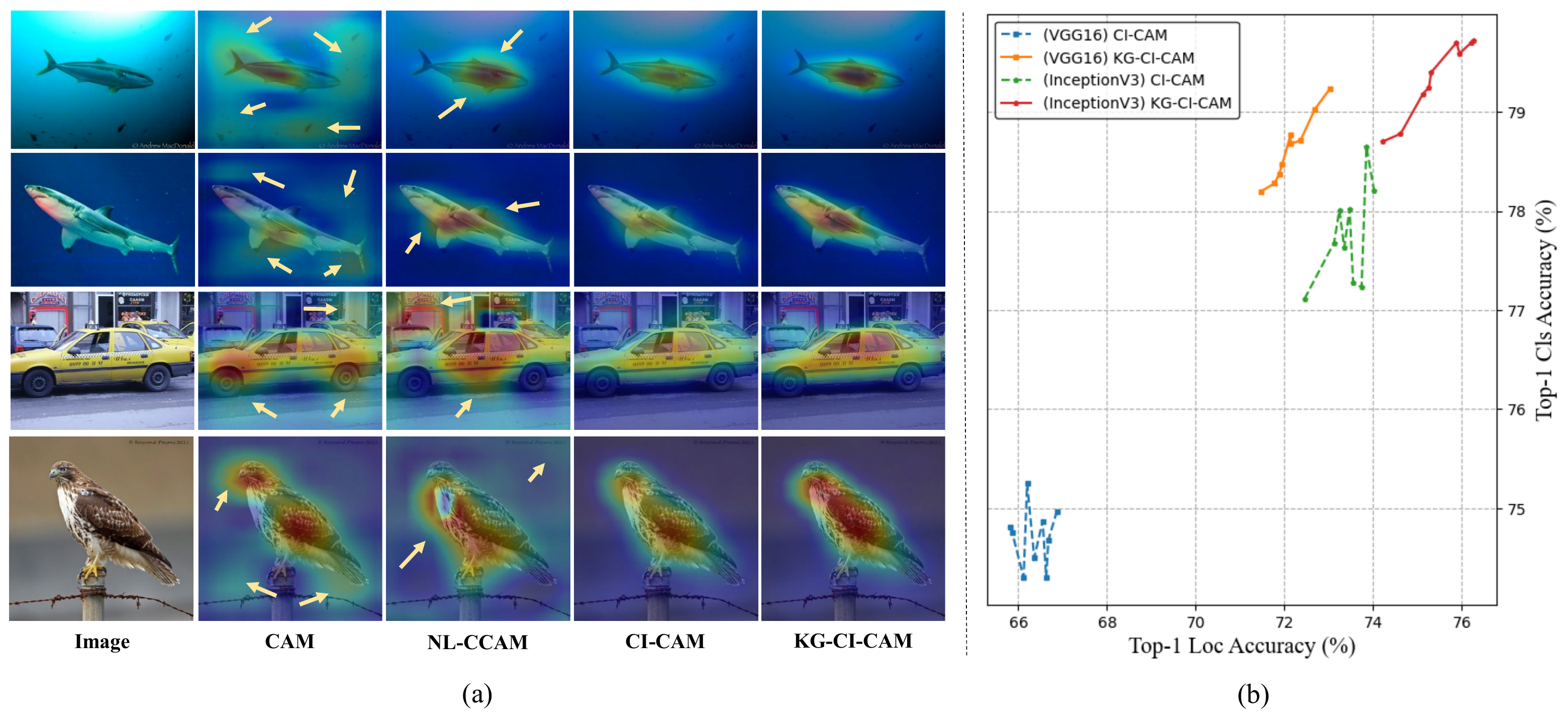}
   \caption{(a) Visualization comparison between vanilla CAM, NL-CCAM, CI-CAM, and KG-CI-CAM. The yellow arrows indicate the regions suffer from entangled contexts. (b) The classification-localization dilemma faced by CI-CAM, where the classification and localization suffer from a performance gap and can not achieve their highest accuracy simultaneously.}
   \label{comparison_graph}
 \end{figure*}

Recently, the techniques based on deep convolutional neural networks (DCNNs)~\cite{simonyan2014very,szegedy2015going,he2016deep,luo2021category,luo2018macro} promote the localization performance to a new level. However, this performance promotion is at the price of huge amounts of fine-grained human annotations\cite{luo2019taking, luo2020ASM, shao2022active}. To alleviate such a heavy burden, weakly-supervised object localization (WSOL) has been proposed by only resorting to image-level labels. To capitalize on the image-level labels, the existing studies~\cite{selvaraju2017grad, diba2017weakly, kim2017two, wei2018ts2c, shao2022deep, wang2022looking, shao2021improving, shao2023mitigating} follow the Class Activation Mapping (CAM) approach~\cite{zhou2016learning} to generate class activation maps first and then segment the highest activation area for a coarse localization. 

Albeit, CAM is initially designed for the classification task and tends to focus only on the most discriminative feature to increase its classification accuracy. To deal with this issue, recent prevailing works~\cite{diba2017weakly, kim2017two, wei2018ts2c, zhang2018adversarial, mai2020erasing, yang2020combinational} endeavor to perceive the whole objects instead of the shrunken and sparse ``discriminative regions''. On one hand, they refine the network structure to make the detector more tailored for object localization in a weak supervision setting. For example, some methods~\cite{diba2017weakly, wei2018ts2c} use a three-stage structure to continuously optimize the prediction results by training the current stage using the output of the previous stages as supervision. Other works~\cite{kim2017two, zhang2018adversarial, mai2020erasing} leverage two parallel branches where the first branch is designed for digging out the most discriminative small regions while the second one is responsible for detecting the less discriminative large regions. On the other hand, they also make full use of the image information to improve the prediction results. For instance, TS$^2$C~\cite{wei2018ts2c} and NL-CCAM~\cite{yang2020combinational} utilize the contextual information of surrounding pixels and the activation maps of low probability class, respectively.

Although the works along the vein of CAM have led to some impressive early results, we have discovered multi-faceted problems that remain unexplored by those prior arts, which emerged from our experimental analysis. \emph{Firstly}, vanilla CAM mechanism is incapable of reasoning about the co-occurrence confounder (\eg, \textit{fish} and \textit{water}) in the images, which makes the model inspection hard to distinguish between the object and context and causes biased activation maps, as shown in Figure~\ref{comparison_graph} (a). We dub this problem as ``entangled context'' for convenience. \emph{Secondly}, we noticed that the classification and localization always suffer from a performance gap and can not reach their highest accuracy simultaneously in previous CAM-based methods. Taking CI-CAM~\cite{shao2021improving} as an example, we set up multiple independent experiments using different training settings, however, none of the results can achieve the best classification and localization performance at the same time, as reported in Figure~\ref{comparison_graph} (b). We summarise this issue as a classification-localization dilemma (``C-L dilemma'' for short). We argue that these two problems severely hinder the WSOL performance and heretofore yet to be well studied, despite the existence of a vast body of WSOL literature~\cite{zhou2016learning, diba2017weakly, zhang2018adversarial, zhang2020inter, mai2020erasing, wu2022background}. 

Taking one step further into these two issues, we found the internal reasons behind them are different. The ``entangled context'' derives from the fact that objects usually co-occur with a certain context background, \eg, the most ``\textit{fish}'' appears concurrently with ``\textit{water}'' in the images. Consequently, these two concepts would be inevitably entangled and a classification model would wrongly generate an ambiguous boundary between ``\textit{fish}'' and ``\textit{water}'' with image-level supervision. In contrast to the vanilla CAM which yields a relatively shrunken bounding box on the small discriminative region, we notice that the ``entangled context'' problem would cause a biased expanding bounding box that includes the wrongly entangled background, which impairs the localization accuracy in terms of the object range. While the ``C-L dilemma'' is derived from a hallmark of a classification network that prefers the discriminative region information (\eg, the head of an animal) over the integral profile of an object to better distinguish various categories. Consequently, pooling layers are extensively utilized in classification networks to focus on these highly representative features. However, these pooling layers inevitably lead to the discarding of unrepresentative information, which is critical information for locating the entire object instance. Forcing a classification model to pay more attention to the unrepresentative area (\eg, the fur of an animal) conceding to the integral contour perception would inevitably cause a biased categorical prediction, and vise versa for a localization model. Prior approaches sidestep such a dilemma by simply trading off the classification and localization performances, \ie, choosing a mutually acceptable model or only reporting localization results~\cite{pan2021unveiling, wu2022background, kim2022bridging, xu2022cream}. However, the intrinsic issue behind this dilemma remains under-explored.

In this paper, we propose a knowledge-guided causal intervention method to solve both ``entangled context'' and ``C-L dilemma'' problems in one go, dubbed KG-CI-CAM. Specifically, we first explore the causalities among image features, contexts, and labels by establishing a structural causal model (SCM)~\cite{pearl2016causal} and pinpoint the context as a confounder as shown in Figure~\ref{structural_causal_model_graph}. According to the causal analysis, we propose a causal context pool to eliminate the biased co-occurrence in the class activation maps to solve the ``entangled context'' problem as shown in Figure~\ref{network_architecture_graph}. Based on disentangled object features from the wrongly contextual information, we further tackle the ``C-L dilemma'' problem faced by the CAM-based model via a multi-source knowledge guidance framework as shown in Figure~\ref{fig:kd}. The classification source aims to provide good logit knowledge and the localization source is responsible for providing high-quality class activation map knowledge. These two sources guide our model to balance the absorption of classification knowledge and localization knowledge. With these tailored designs, KG-CI-CAM can not only effectively eliminate the spurious correlations between pixels and labels from the ``entangled context'' problem, but also alleviate the ``C-L dilemma'' problem resulting from the interesting bias between the discriminative region information over the integral profile of an object.

It is noteworthy that some pertinent results of this work have been published in an earlier version~\cite{shao2021improving}. This paper goes beyond~\cite{shao2021improving} by contributing to introducing a novel knowledge-guided causal intervention method to further boost the WSOL by solving the notorious but under-explored ``entangled context'' and ``C-L dilemma'' problems in one go. We also make significant improvements in experimental validation and analysis compared to the early version.

The main contributions can be summarized as follows:

\begin{itemize}
   \item We are among the pioneers to concern and reveal the inevitable ``entangled context'' and ``C-L dilemma'' issues of WSOL that remain unexplored by prevailing efforts.
   \item We propose a novel knowledge-guided causal intervention method, dubbed KG-CI-CAM, to address the confounding context and guide the model absorption of classification and localization knowledge in one go.
   \item Extensive experiments show that both ``entangled context'' and ``C-L dilemma'' problems are effectively solved by our proposed method.
\end{itemize}

\section{Related Work}
\subsection{Weakly-supervised Object Localization}
Since CAM~\cite{zhou2016learning} often exhibits bias towards the most discriminative object part rather than the entire object, current research primarily centers around enhancing object localization accuracy. These methods can be broadly classified into two categories: expanding proposal regions and removing discriminative regions. The first one is to appropriately enlarge the size of the initial prediction box~\cite{diba2017weakly, wei2018ts2c}. WCCN~\cite{diba2017weakly} employs a three-stage cascaded network, progressively enlarging and refining proposal outputs. TS$^2$C~\cite{wei2018ts2c} selects the final box by comparing mean pixel confidence values of the initial prediction region and its surroundings, considering a gap between these means. The other is to detect larger regions after removing the most discriminative part~\cite{kim2017two, zhang2018adversarial, choe2019attention, mai2020erasing}. TP-WSL~\cite{kim2017two} identifies the most discriminative region, erasing it from conv5-3 feature maps in a second network. ACoL~\cite{zhang2018adversarial} utilizes masked feature maps by erasing the most discriminative region found by the first classifier as input for the second classifier. ADL~\cite{choe2019attention} stochastically generates an erased mask or importance map as an attention map projected onto image feature maps. MEIL~\cite{mai2020erasing} employs adversarial erasing, simultaneously computing erased and unerased branches through a shared classifier.

While the above methods primarily address localization issues stemming from the most discriminative object parts, they overlook the challenge of fuzzy boundaries between objects and co-occurring contextual backgrounds. For instance, if ``fish'' frequently appears alongside ``water'' in images, these concepts may become intertwined, leading to ambiguous boundaries under image-level supervision alone.

\subsection{Causal Inference}
Structural Causal Model (SCM)\cite{pearl2016causal}, a key tool in causal inference\cite{yue2020interventional, zhang2020causal, tang2020long, hu2022causal}, employs a directed graph where nodes represent model participants and links signify causal relationships~\cite{zhang2020causal}. Zhang \etal~\cite{zhang2020causal} deeply analyze causal connections among image features, contexts, and class labels using SCM~\cite{didelez2001judea, pearl2016causal}, proposing Context Adjustment (CONTA) which sets a new state-of-the-art in weakly-supervised semantic segmentation. Yue \etal~\cite{yue2020interventional} employ causal intervention in few-shot learning, revealing pre-trained knowledge as a limiting confounder, leading to their novel paradigm: Interventional Few-Shot Learning (IFSL). Tang \etal~\cite{tang2020long} demonstrate that SGD momentum is essentially a confounder through employing SCM in long-tailed classification.

In this paper, our approach diverges from the alternating training strategy between the WSOL model and semantic segmentation model in CONTA~\cite{zhang2020causal}, as we propose an integrated model embedding causal inference into the WSOL pipeline. This integrated model uses a causal context pool for enhancing feature boundary clarity, which is detailed in \S\ref{sec:causal_intervention}.

\subsection{Knowledge Distillation}
Knowledge distillation~\cite{hinton2015distilling, gou2021knowledge} aims to transfer insights from a large teacher model to a compact student model~\cite{luo2024LDMC}, which can be split into knowledge and distillation. Knowledge encompasses response-based~\cite{hinton2015distilling, wang2017knowledge} and feature-based~\cite{zheng2022localization, li2024bgae, li2023multi} aspects. Response-based knowledge pertains to prediction logits, effective for simple classification distillation. However, for intricate distillation, relying solely on teacher logits is inadequate, prompting the exploration of feature-based knowledge to enhance distillation efficacy. Distillation, the mode of knowledge transfer, includes offline~\cite{huang2017like, heo2019knowledge}, online~\cite{chen2020online, wu2021peer}, and self-distillation~\cite{lan2018self, yun2020regularizing}. To facilitate smooth knowledge absorption by student models, a ``soft target'' approach is adopted, involving a ``softmax'' with a temperature $T$ applied to teacher knowledge as additional supervision for students.

In this study, we design a multi-source knowledge guidance framework to guide the absorption of classification and localization knowledge during model training, elaborated in \S\ref{sec:multi_teacher_distillation}.

\addtolength{\tabcolsep}{-4.5pt} 
\begin{table*}[t]
   \centering
   \caption{List of symbols and their corresponding descriptions.}
   \label{tab:symbols}
   \begin{tabular}{|l|l|l|l|}
   \hline

   \hline                                            
   Symbols & Descriptions & Symbols & Descriptions \\ 
   \hline
   \multicolumn{4}{|c|}{Causal Intervention for ``Entangled Context'' Issue} \\  \hline  
   $I$  &   input image  & $H$   &  localization map \\
   $X$  &  the feature maps of an image & $X^e$  &  the enhanced feature maps after causal intervention \\
   $S$ &  the prediction score of classification & $S^e$   & the enhanced prediction score of classification after causal intervention \\
   $M$  &   class activation maps & $M^e$  &  the enhanced class activation maps after causal intervention \\
   $P()$ &   predicted probability & $C$ &    confounding context \\
   $Y$   &  image label & $V$  &   an image-specific representation created using contextual template  \\
   $Q$ &   causal context pool & $d, h, w, n$  &  the channel count, height, width of feature maps, the number of classes \\
   \hline
   \multicolumn{4}{|c|}{Knowledge Guidance for ``C-L Dilemma'' Issue} \\  \hline  
   $Z$   &  the output logits of CI-CAM & $\tilde{Z}$  &  the output logits of the classification expert \\
   $A$  &   the class activation maps of CI-CAM & $\tilde{A}$ &  the class activation maps of the localization expert \\
   $T^{cls}$  & the distillation temperature of logits  & $T^{loc}$ & the distillation temperature of class activation maps \\
   $M^{fore}$ & a binary foreground mask &  $M^{back}$ & a binary background mask \\
   $I^{fore}$ & a foreground image  & $I^{back}$  & a background image \\
   $S^{fore}$ & the prediction score of classification in an foreground image & $S^{back}$ & the prediction score of classification in an background image \\
   \hline

   \hline
   \end{tabular}
\end{table*}
\addtolength{\tabcolsep}{4.5pt} 

\section{Methodology}
We begin by presenting the baseline approach for WSOL in \S\ref{sec:baseline}. Subsequently, our focus shifts to addressing the ``entangled context'' challenge, leading to the introduction of our causal intervention method (CI-CAM) in \S\ref{sec:causal_intervention}. Building upon the causal model, we then address the ``C-L dilemma'' and formulate the multi-source knowledge guidance framework (KG-CI-CAM) in \S\ref{sec:multi_teacher_distillation} and model training in \S\ref{sec:model_training}. Table~\ref{tab:symbols} presents the symbols and their descriptions.

\subsection{Combinational Class Activation Mapping}
\label{sec:baseline}
Given an image $I$, it undergoes processing through a fully convolutional backbone to yield its feature maps $X \in \mathbb{R}^{d \times h \times w}$, where $d$, $h$, and $w$ respectively signifies the channel count, height, and width of feature maps. Subsequently, $X$ is directed through a global average pooling (GAP) layer, followed by a classifier with a fully connected layer. Utilizing a softmax layer atop this classifier computes prediction score $S=\{s_1, s_2, \ldots, s_n\}$ for classification. The classifier's weight matrix is denoted as $W \in \mathbb{R}^{n \times d}$, with $n$ representing the number of image classes. Consequently, the activation map $M_i$ corresponding to class $i$ within the class activation maps $M \in \mathbb{R}^{n \times h \times w}$, as proposed in~\cite{zhou2016learning}, is expressed as follows:
\begin{equation}
   \begin{aligned}
      M_i = \sum_{k}^{d} {W_{i, k} \cdot X_k}, i \in \{1,2, \ldots, n\}.
   \end{aligned}
   \label{eq:cam}
\end{equation}

Yang \etal~\cite{yang2020combinational} assert that using solely the activation map of the class with the highest probability for segmenting object boxes can be problematic, often favoring overly small regions or inadvertently highlighting background areas. Thus, they introduce NL-CCAM~\cite{yang2020combinational}, which amalgamates all class activation maps into a localization map $H$, designed as follows:
\begin{equation}
   \begin{aligned}
      H = \sum_{i}^n \gamma(i)\cdot M_i,
   \end{aligned}
   \label{eq:ccam}
\end{equation}
where $\gamma(i)$ corresponds to a weight value linked to the rank of class $i$ among all class probabilities. Ultimately, they apply a threshold, as proposed in~\cite{zhou2016learning}, to segment $H$, thereby generating a bounding box for object localization.

Our method is based on NL-CCAM~\cite{yang2020combinational} but introduces substantial improvements. We not only equip the baseline network with the ability of causal inference to tackle the ``entangled context'' problem but also address the ``C-L dilemma'' problem suffered from the traditional CAM-based models.

\subsection{Causal Intervention for ``Entangled Context'' Issue}
\label{sec:causal_intervention}
We first elucidate the detrimental impact of the confounding context through the application of a structural causal model~\cite{pearl2016causal} in \S\ref{sec:structural_causal_model}. Subsequently, we employ causal intervention to theoretically address the issue of ``entangled context'' using backdoor adjustment~\cite{pearl2016causal} in \S\ref{sec:theoretical_analysis}. Lastly, we integrate the causal intervention into our model in \S\ref{sec:stu_network_structure}.

\subsubsection{Structural Causal Model}
\label{sec:structural_causal_model}
\begin{figure}[t]
   \centering
   \includegraphics[width=1.0\linewidth]{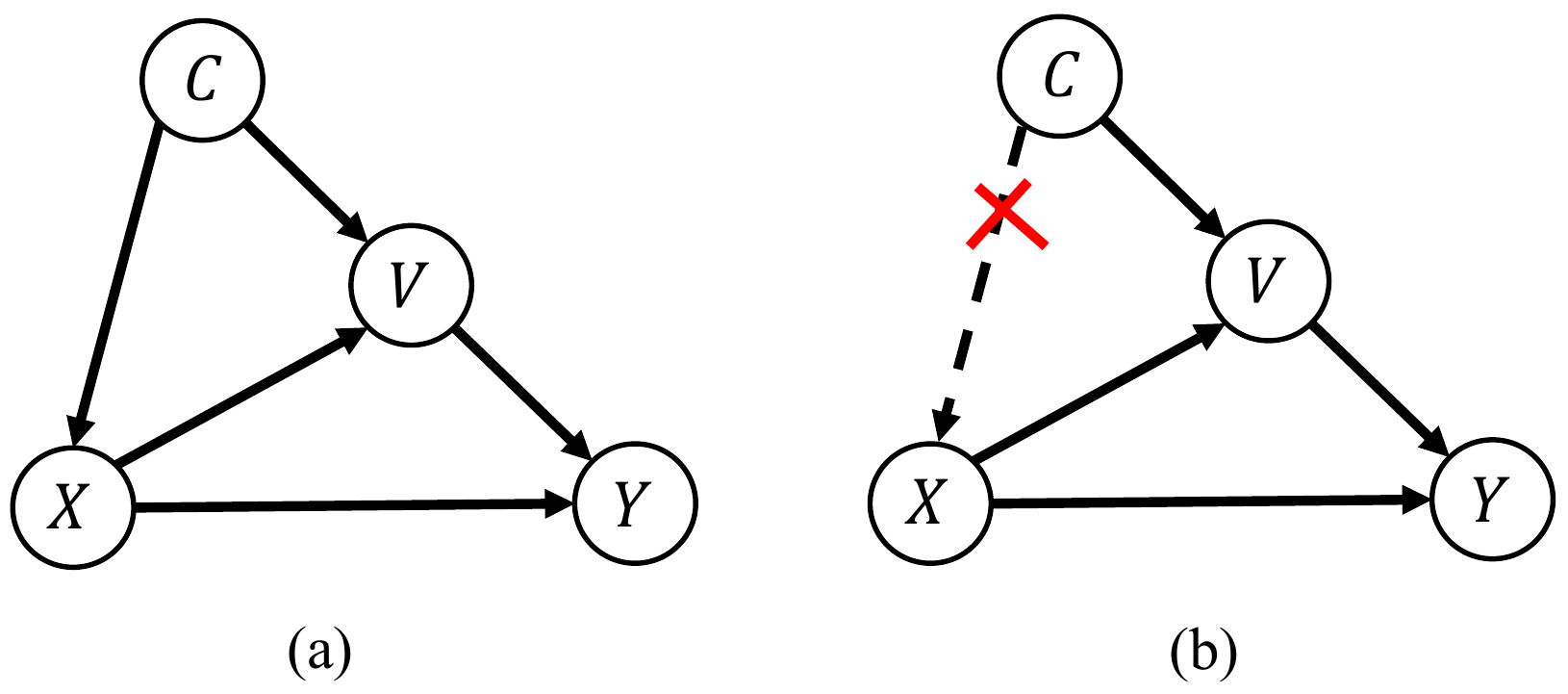}
   \caption{(a) Building the structural causal model (SCM) in WSOL. (b) Cutting off the confounding effect of $C \rightarrow X$ in WSOL. $X$: feature maps. $C$: confounding context. $V$: image representation. $Y$: image label.}
   \label{structural_causal_model_graph}
 \end{figure}

Drawing inspiration from CONTA~\cite{zhang2020causal}, we employ a structural causal model (SCM)\cite{pearl2016causal} to analyze the causal relationships among image feature $X$, confounding context $C$, and image-level label $Y$. The directed link presented in Figure~\ref{structural_causal_model_graph} (a) signifies the causal connection between the respective nodes: cause $\rightarrow$ effect~\cite{zhang2020causal}.

$\bm{C \rightarrow X}$: This link signifies that the feature maps $X$ are generated by the backbone in the presence of the context $C$. Although the confounding context $C$ aids in establishing a stronger connection between image features $X$ and labels $Y$ through the prediction probability $P(Y|X)$, \textit{e.g.}, it is likely a ``fish'' when perceiving a ``water'' region, $P(Y|X)$ mistakenly associates non-causal but positively correlated pixels to labels, \textit{e.g.}, the ``water'' region wrongly belongs to ``fish''. This is a vital reason for the inaccurate localization in WSOL. Fortunately, as we will discuss later in \S\ref{sec:stu_network_structure}, we can circumvent this issue by employing a causal context pool within the framework of causal intervention.
 
$\bm{C \rightarrow V \leftarrow X}$: $V$ represents an image-specific representation created using contextual templates from $C$\cite{zhang2020causal}. For instance, $V$ offers insights into the shape and location of a ``fish'' (foreground) within a scene (background)\cite{zhang2020causal}. In this paper, $V$ denotes the activation map of the highest probability class in the CAM 1 module, as depicted in Figure~\ref{network_architecture_graph}.
 
$\bm{X \rightarrow Y \leftarrow V}$: These connections denote both image feature $X$ and image representation $V$ jointly influence image label $Y$. We argue the shape and location information of the object instance contained in image representation $V$ directly impact the image label $Y$. Consequently, although $V$ is not an input factor in the WSOL model, it remains influential~\cite{zhang2020causal}.


\subsubsection{Theoretical Analysis}
\label{sec:theoretical_analysis}
To mitigate the confounding influence of $\bm{C \rightarrow X}$, as illustrated in Figure~\ref{structural_causal_model_graph} (b), we draw inspiration from CONTA~\cite{zhang2020causal}. Following the same principle, we employ the backdoor adjustment~\cite{pearl2016causal} to utilize $P(Y|do(X))$ as the novel image-level classifier, which removes the confounder $C$ and pursues the true causality from $X$ to $Y$ shown in Figure~\ref{structural_causal_model_graph} (b). In this way, we can achieve better classification and localization in WSOL. The key idea is to 1) cut off the link $C \rightarrow X$, and 2) stratify $C$ into pieces $C = \{c_1, c_2, \ldots, c_n\} $, and $c_i$ denotes the $i^{th}$ class context. Formally, we have 
\begin{equation}
   \begin{aligned}
      P(Y|do(X)) = \sum_{i}^{n} P(Y|X=x, V=f(x, c_i))\cdot P(c_i),
   \end{aligned}
   \label{eq:p_do_x}
\end{equation}
where $f(X,c)$ abstractly represents that $V$ is formed by the combination of $X$ and $c$, and $n$ is the number of image class. As $C$ does not affect $X$, it guarantees $X$ to have a fair opportunity to incorporate every context $c$ into $Y$'s prediction, subject to a prior $P(c)$. Inspired by CONTA~\cite{zhang2020causal}, we use the Normalized Weighted Geometric Mean~\cite{xu2015show} to optimize Eq.~\eqref{eq:p_do_x} by moving the outer sum $\sum_i^n P(c_i)$ into the feature.
\begin{equation}
   \begin{aligned}
      P(Y|do(X)) \approx P(Y|X=x, V=\sum_i^n f(x, c_i) \cdot P(c_i)).
   \end{aligned}
   \label{eq:p_do_x_nwgm}
\end{equation}
Since the number of samples for each class in the dataset is roughly the same, we set $P(c)$ to uniform $1/n$. After further optimizing Eq.~\eqref{eq:p_do_x_nwgm}, we have
\begin{equation}
   \begin{aligned}
      P(Y|do(X)) \approx P(Y|x \oplus \frac{1}{n} \cdot \sum_i^n f(x, c_i)),
   \end{aligned}
   \label{eq:p_do_x_final}
\end{equation}
where $\oplus$ denotes projection. So far, the ``entangled context'' issue has been transferred into calculating $\sum_i^n f(x, c_i)$. We introduce a causal context pool $Q$ to represent it in \S\ref{sec:stu_network_structure}.

\subsubsection{Causal Network Architecture}
\label{sec:stu_network_structure}
\begin{figure*}[t]
   \centering
   \includegraphics[width=1.0\linewidth]{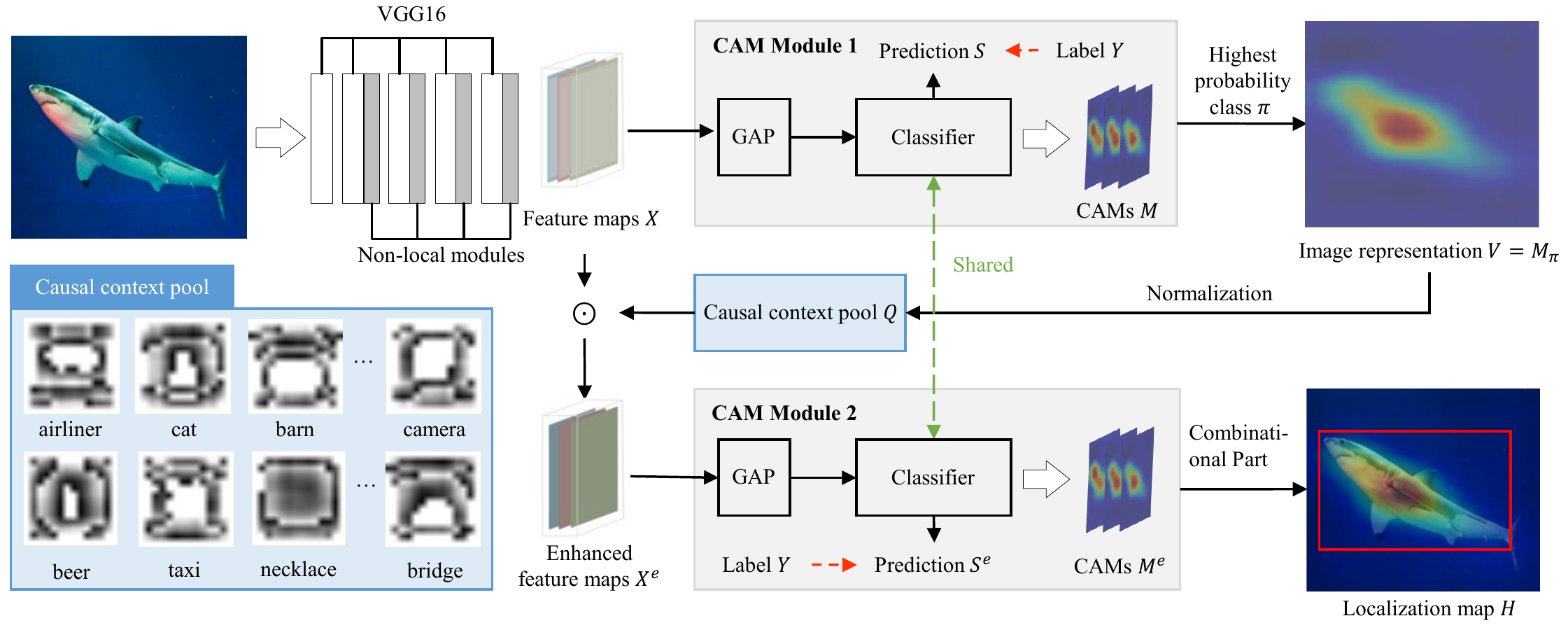}
   \caption{Overview of the proposed causal network architecture: \textbf{CI-CAM}. CI-CAM consists of four parts: a backbone to extract the feature maps, the share-weighted CAM modules to generate class activation maps, a causal context pool to enhance the feature maps by eliminating the negative effect of confounder, and a combinational module to generate the final bounding box.}
   \label{network_architecture_graph}
 \end{figure*}

 We implement a causal model for the ``entangled context'' problem, dubbed CI-CAM, at the core of which is the incorporation of a causal context pool. The primary concept involves aggregating all class contexts, and subsequently re-mapping these contexts onto the feature maps of convolutional layers as detailed in Eq.\eqref{eq:p_do_x_final}. The objective is to establish a pure causality between the cause $X$ and the effect $Y$. Figure~\ref{network_architecture_graph} illustrates the overview of CI-CAM that includes four parts: backbone, CAM module, causal context pool, and combinational part.

 \textbf{Backbone:} Drawing inspiration from the baseline method, we formulate our backbone architecture by seamlessly integrating multiple non-local blocks into both low- and high-level layers of a fully convolutional network. This configuration serves as a feature extractor, operating on RGB images as input and generating position-aware feature maps.
 
 \textbf{CAM Module:} This module comprises a Global Average Pooling (GAP) layer and a classifier with a fully connected layer~\cite{zhou2016learning}. Image feature maps $X$, generated by the backbone, are passed through GAP and the classifier to generate prediction score $S=\{s_1, s_2, \ldots, s_n\}$. The CAM network multiplies the classifier's weight matrix $W$ with $X$ to produce class activation maps $M \in \mathbb{R}^{n \times h \times w}$ as described in Eq.~\eqref{eq:cam}. In our model, we employ two CAM modules with shared weights. The first CAM module is designed to generate initial prediction score $S$ and class activation maps $M$, while the second CAM network is responsible for producing more accurate prediction score $S^e=\{s_1^e, s_2^e, \ldots, s_n^e\}$ and class activation maps $M^e \in \mathbb{R}^{n \times h \times w}$. This refinement in the second CAM module is achieved using the feature maps $X^e \in \mathbb{R}^{d \times h \times w}$, which have been enhanced by the causal context pool as described in Figure~\ref{network_architecture_graph}.
 
 \textbf{Causal Context Pool:} Within the training phase, we maintain a causal context pool denoted as $Q \in \mathbb{R}^{n \times h \times w}$. The pool $Q$ consistently preserves all contextual information maps associated with each class. This accumulation is achieved by storing the activation map of the highest probability class. Subsequently, it deploys these accumulated contexts as attention mechanisms onto the feature maps of the backbone layer, resulting in feature maps that have been augmented. The rationale behind the use of a causal context pool extends beyond mitigating the adverse influence of entangled context on image feature maps; it also serves to emphasize the positive regions within the image feature maps, thereby enhancing localization performance.

\textbf{Combinational Part:} The input to the combinational part comprises the class activation maps $M^e$ produced by the second CAM module. The resulting output is a localization map $H \in \mathbb{R}^{h \times w}$, which is computed using Eq.\eqref{eq:ccam}. 

\subsubsection{Data Flow and Training Objective}
With all the key modules presented above, let's provide a concise overview of the data flow within our network. When provided with an image $I$, our initial step is to pass it through the backbone, generating feature maps $X$. These feature maps $X$ are subsequently input into two parallel CAM modules. The first CAM module is responsible for producing initial prediction score $S$ and class activation maps $M$. Following this, the causal context pool $Q$ undergoes an update process, achieved by amalgamating the activation map $M_{\pi}$ (presented as $V$ in Figure~\ref{structural_causal_model_graph}), as outlined below:
\begin{equation}
   \begin{aligned}
      Q_\pi = bn(Q_\pi + \lambda \cdot bn(M_\pi)),
   \end{aligned}
   \label{eq:update_causal_context_pool_1}
\end{equation}
where $\pi = argmax(\{s_1, s_2, \ldots, s_n\})$, $Q_\pi$ signifies the contextual map of class $\pi$, $\lambda$ stands for the update rate, and $bn$ denotes batch normalization. The responsibility of the second CAM module lies in generating prediction score $S^e$ with higher precision, and refining class activation maps $M^e$. The input of the second branch is enhanced feature maps $X^e$ projected by the causal context pool $Q$. More concretely, the feature enhancement can be calculated as follows:
\begin{equation}
   \begin{aligned}
      X^e = X + X\odot Conv_{1 \times 1}(Q_\pi),
   \end{aligned}
\end{equation}
where the symbol $\odot$ represents the matrix dot product. Within the combinational part, we aim to construct a localization map $H \in \mathbb{R}^{h \times w}$ using the refined class activation maps $M^e$. Subsequently, we utilize a straightforward thresholding technique proposed by~\cite{zhou2016learning} to generate a bounding box $B$ based on $H$. Ultimately, the bounding box $B$ and the prediction score $S^e$ together constitute the final prediction.

During the training phase, our causal network is designed to minimize image classification losses for both CAM modules, utilizing the following loss function denoted as $\mathcal{L}^{causal}$.
\begin{equation}
   \begin{aligned}
      \mathcal{L}^{causal} = \rho \cdot cross\_entropy(S, Y) + cross\_entropy(S^e, Y),
   \end{aligned}
   \label{eq:causal_loss}
\end{equation}
where $Y$ is the ground-truth label of an image. If $\mathcal{L}^{causal}$ is used in Eq.~\ref{eq:cls_tch_loss} and Eq.~\ref{eq:loc_tch_loss}, $\rho = 1$; otherwise, $\rho = 0$.

\subsection{Knowledge Guidance for ``C-L Dilemma'' Issue}
\label{sec:multi_teacher_distillation}

\begin{figure*}[t]
   \begin{center}
      \includegraphics[width=1.0\linewidth]{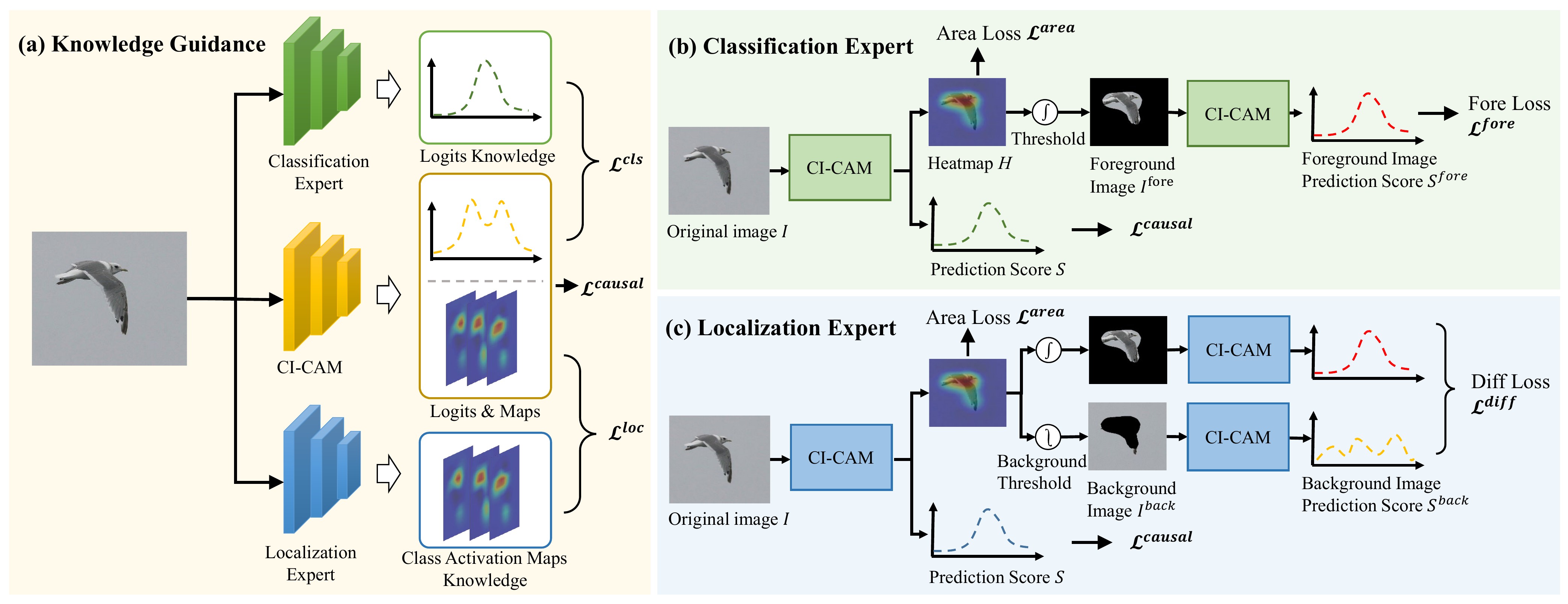}
   \end{center}
   \caption{(a) Overview of the multi-source knowledge guidance framework. (b) Overview of the classification expert network in which two green CI-CAM models are share-weighted with each other. (c) Overview of the localization expert network, featuring three blue CI-CAM models, all of which are identical.} 
   \label{fig:kd}
\end{figure*}
 
\subsubsection{Multi-source Knowledge Guidance Framework}
\label{sec:distillation_framework}
To address the challenge posed by the ``C-L dilemma'' and simultaneously optimize both classification and localization performance, we have developed a multi-source knowledge guidance framework. This framework incorporates classification knowledge and localization knowledge~\cite{liu2021combining, von2021informed} to guide the learning process of our CI-CAM model shown in Figure~\ref{fig:kd} (a).

Initially, our focus centers on obtaining high-quality classification expertise from a specialist in the field. This expertise is utilized to supervise the CI-CAM model and enhance its classification performance. To achieve this, we have designed a logits guidance loss function $\mathcal{L}^{cls}$, which establishes a connection between the classification expert and the CI-CAM model, operating as follows:
\begin{equation}
   \begin{aligned}
      \mathcal{L}^{cls} = KL(soft(Z, T^{cls}), soft(\tilde{Z}, T^{cls})),
   \end{aligned}
\end{equation}
where $soft(Z, T)=\frac{\exp(Z_i/T)}{\sum_{j=1}^{n} \exp(Z_j/T)}$. $KL$ represents the Kullback-Leibler divergence function. $Z$ and $\tilde{Z}$ correspond to the output logits of the CI-CAM model and the classification expert, respectively. $n$ and $T^{cls}$ stand for the number of classes and the temperature used for guiding logits, respectively.

Moving on, our second objective is to precisely localize the entire object rather than focusing solely on its most distinctive part. To accomplish this, we harness the localization knowledge provided by an expert in the field. Specifically, we have developed an activation guidance loss function $\mathcal{L}^{loc}$, which establishes a link between the localization expert and the CI-CAM model, and operates as follows:
\begin{equation}
   \begin{aligned}
      \mathcal{L}^{loc} = MSE(soft(A, T^{loc}), soft(\tilde{A}, T^{loc})),
   \end{aligned}
\end{equation}
where $MSE$ is the Mean Squared Error function. $A$ and $\tilde{A}$ respectively denote the class activation maps of the CI-CAM model and the localization expert. The variable $T^{loc}$ refers to the temperature used for guiding activation.

Concluding our approach, the comprehensive loss function $\mathcal{L}^{guidance}$ for the multi-source knowledge guidance framework depicted in Figure~\ref{fig:kd} (a) can be expressed as follows:
\begin{equation}
   \label{eq:guidance}
   \begin{aligned}
      \mathcal{L}^{guidance} = \alpha \cdot (\mathcal{L}^{cls} + \mathcal{L}^{loc}) + (1-\alpha)\cdot \mathcal{L}^{causal},
   \end{aligned}
\end{equation}
where $\alpha$ is the guiding hyper-parameter.

\subsubsection{Classification Expert}
\label{sec:classification_teacher}
To prevent the degradation of knowledge transfer from the model capacity gap between CI-CAM and expert model~\cite{mirzadeh2020improved, gao2021residual}, we construct our classification expert by adopting CI-CAM model, which is shown in Figure~\ref{fig:kd} (b). Concretely, given an input image $I$, the process begins with passing it through the model to generate the original image prediction score $S$ and the localization map $H$. Subsequently, we generate a binary foreground mask $M^{fore}$ through segmentation of the localization map $H$ utilizing a threshold. This mask $M^{fore}$ is then employed to create a foreground image $I^{fore}$, which is obtained by overlaying the mask onto the original image $I$. Finally, the foreground image $I^{fore}$ is once again inputted into the model to generate the foreground image prediction score $S^{fore}$.

To compel the classification expert to focus on the most discriminative foreground information while disregarding the background information, we introduce two extra losses: the Area Loss~\cite{xie2021online, meng2021foreground} and the Fore Loss. The purpose of these losses is to enhance the prominence of foreground regions while suppressing background areas. On the one hand, the Area Loss focuses on diminishing the activation values attributed to both foreground and background areas within the localization map $H$. On the other hand, the Fore Loss is tasked with activating the regions corresponding to the object within the image by means of classifying the foreground image. For clarity, the Area Loss $\mathcal{L}^{area}$ and the Fore Loss $\mathcal{L}^{fore}$ are defined as follows:
\begin{equation}
   \label{eq:area_loss}
   \begin{aligned}
      \mathcal{L}^{area} = \frac{1}{h\cdot w} \sum_{i=1}^{h} \sum_{j=1}^{w} H_{i, j},
   \end{aligned}
\end{equation}
\begin{equation}
   \label{eq:fore_loss}
   \begin{aligned}
      \mathcal{L}^{fore} = cross\_entropy(S^{fore}, Y),
   \end{aligned}
\end{equation}
where $Y$ is the ground-truth label of an image. As a consequence, the total training loss function of our classification expert is designed as follows:
\begin{equation}
   \label{eq:cls_tch_loss}
   \begin{aligned}
      \mathcal{L}^{cls\_expert} = \mathcal{L}^{causal} + \mu \cdot \mathcal{L}^{fore} + \eta \cdot \mathcal{L}^{area},
   \end{aligned}
\end{equation}
where $\mu$ and $\eta$ are the hyper-parameters. In Figure~\ref{fig:teacher_add} (a), the comparative results between CI-CAM and the classification expert are presented, thereby confirming the superior classification performance attainable through our classification expert.

\subsubsection{Localization Expert}
\label{sec:localization_teacher}
\begin{figure}[t]
   \centering
   \includegraphics[width=0.8\linewidth]{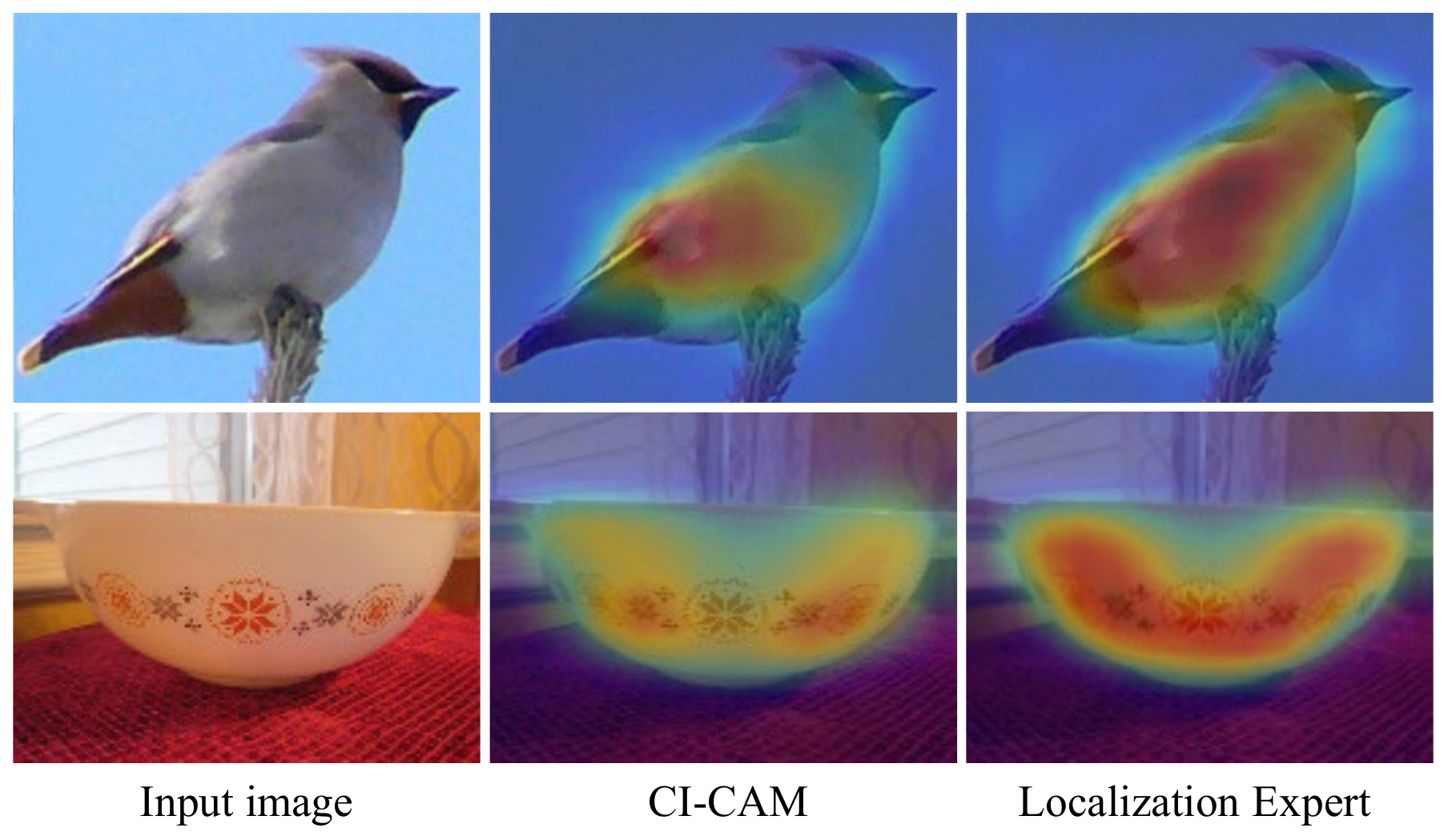}
   \caption{Comparison between CI-CAM and localization expert.}
   \label{fig:stu_loc}
 \end{figure}

Similar to the classification expert, we construct our localization expert by adopting the CI-CAM model as shown in Figure~\ref{fig:kd} (c). To delve further into specifics, given an input image $I$, the initial step involves passing it through the model to generate both the original image prediction score $S$ and the localization map $H$. Subsequently, we proceed to obtain two distinct masks through the segmentation of the localization map $H$ using a thresholding approach: the binary foreground mask $M^{fore}$ and the binary background mask $M^{back}$. Moving forward, we employ these masks to generate the foreground image $I^{fore}$ and the background image $I^{back}$ by projecting the respective masks onto the original image $I$. Lastly, both the foreground image $I^{fore}$ and the background image $I^{back}$ are reintroduced into the model to produce the corresponding prediction scores: $S^{fore}$ for the foreground image and $S^{back}$ for the background image. 

\begin{figure*}[t]
   \centering
   \includegraphics[width=0.8\linewidth]{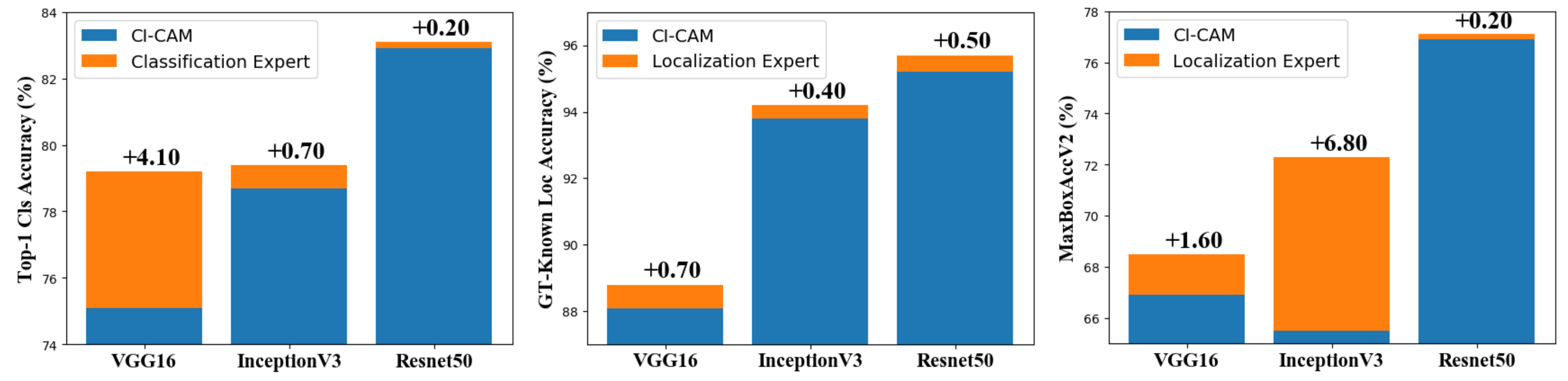}
   \caption{Quantitative comparison between CI-CAM and experts. Orange areas are the improvements from experts.}
   \label{fig:teacher_add}
 \end{figure*}

To make the foreground image contain the integral object as much as possible while reducing the object information in the background area, we design an extra difference loss (Diff Loss) $\mathcal{L}^{diff}$, which is given as follows:
\begin{equation}
   \label{eq:diff_loss}
   \begin{aligned}
      \mathcal{L}^{diff} = & cross\_entropy(S^{fore}-S^{back}, Y) \\
       & + cross\_entropy(\frac{S + (S^{fore}-S^{back})}{2}, Y),
   \end{aligned}
\end{equation}
where $Y$ and $S$ are the ground-truth labels and the original image prediction score, respectively. Thus,  the total training loss function of our localization expert is designed as follows:
\begin{equation}
   \label{eq:loc_tch_loss}
   \begin{aligned}
      \mathcal{L}^{loc\_expert} & = \mathcal{L}^{causal} + \beta \cdot \mathcal{L}^{diff} + \delta \cdot \mathcal{L}^{area},
   \end{aligned}
\end{equation}
where $\beta$ and $\delta$ are the hyper-parameters. The comparison results between CI-CAM and the localization expert, substantiating the localization expert's capacity to generate high-quality localization maps and attain superior localization performance, are visually illustrated in Figure~\ref{fig:stu_loc} and Figure~\ref{fig:teacher_add} (b and c).

\subsection{Model Training}
\label{sec:model_training}
To simultaneously address the ``entangled context'' and ``C-L dilemma'' challenges, our model's training phase primarily encompasses two key steps. The initial step involves training classification and localization experts, while the subsequent step entails training KG-CI-CAM, guided by the expertise acquired from these trained experts.

More specifically, we begin by training our classification and localization experts, employing Eq.\ref{eq:cls_tch_loss} and Eq.\ref{eq:loc_tch_loss} respectively. It's important to note that our experts are all built upon the CI-CAM model, endowing them with the capability to effectively tackle the ``entangled context'' issue. Subsequently, we feed the training images to the previously trained classification and localization experts, thus acquiring valuable classification and localization knowledge. Finally, we leverage Eq.~\ref{eq:guidance} to train our detection model, which serves to address the ``C-L dilemma'' problem.

\section{Experiments}
\subsection{Datasets and Evaluation Metrics}
\textbf{Datasets.} The evaluation of the proposed KG-CI-CAM method was conducted using two publicly available datasets: CUB-200-2011~\cite{wah2011caltech} and ILSVRC 2016~\cite{russakovsky2015imagenet}. \textbf{CUB-200-2011} is an extension of the Caltech-UCSD Birds 200 (CUB-200) dataset~\cite{welinder2010caltech}, encompassing a collection of $200$ bird species primarily designed to explore subordinate categorization. CUB-200-2011 comprises $5,994$ images in the training set and $5,794$ images in the test set. \textbf{ILSVRC 2016} serves as the dataset initially established for the ImageNet Large Scale Visual Recognition Challenge (ILSVRC)~\cite{russakovsky2015imagenet}. It boasts a comprehensive collection of $1.2$ million images across $1,000$ categories within the training set, with $50,000$ images in the validation set and $100,000$ images forming the test set. 

\textbf{Evaluation Metrics.} We utilize Accuracy and MaxBoxAccV2~\cite{choe2020evaluating} as our primary evaluation metrics. \textbf{Accuracy} encompasses Top-1 classification accuracy (Top-1 Cls), Top-1 localization accuracy (Top-1 Loc), and GT-known localization accuracy (GT-known). Top-1 Cls signifies the correctness of the highest prediction score. Top-1 Loc evaluates the accuracy of both category prediction and box positioning. GT-known focuses exclusively on the accuracy of box positioning prediction. \textbf{MaxBoxAccV2}, introduced in WSOLEvaluation~\cite{choe2020evaluating}, addresses the impact of varying segmentation thresholds on model localization performance. Besides, it offers a more comprehensive evaluation by averaging performance across three Intersection over Union (IoU) thresholds (e.g., 0.3, 0.5, 0.7) to cater to different localization precision requirements.


\subsection{Implementation Details}
We employ pre-trained VGG16~\cite{simonyan2014very}, InceptionV3~\cite{szegedy2016rethinking}, and Resnet50~\cite{he2016deep} models from ImageNet~\cite{russakovsky2015imagenet} as our backbone architectures. Additionally, we insert three non-local blocks into each backbone, initializing these newly added blocks randomly except for the batch normalization layers, which are initialized to zero~\cite{yang2020combinational}. To augment the training data, we apply RandAugment~\cite{szegedy2016rethinking} on the CUB-200-2011~\cite{wah2011caltech} dataset. During the training of experts and KG-CI-CAM, we utilize the Adam optimizer~\cite{kingma2014adam} with $\beta_1 = 0.9$ and $\beta_2 = 0.99$ for both the CUB-200-2011~\cite{wah2011caltech} and ILSVRC 2016~\cite{russakovsky2015imagenet} datasets.

\textbf{Training Classification Expert}: On the CUB, we train our classification experts with a batch size of $bz=6$, an update rate $\lambda = 0.01$, and for $100$ epochs. The specific hyperparameters are as follows:
1) For VGG16: learning rate $lr=0.0005$, $\mu=1.0$, $\eta=0.04$.
2) For InceptionV3: $lr=0.0001$, $\mu=0.6$, $\eta=0.0$.
3) For Resnet50: $lr=0.000125$, $\mu=1.0$, $\eta=0.0$.
On the ILSVRC, we train our classification experts with an update rate $\lambda = 0.001$ for $20$ epochs. The specific hyperparameters are as follows:
1) For VGG16: batch size $bz=42$, learning rate $lr=2.5e-6$, $\mu=1.0$, $\eta=0.04$.
2) For InceptionV3: $bz=72$, $lr=0.002$, $\mu=1.0$, $\eta=0.0$.
3) For Resnet50: $bz=32$, $lr=0.00075$, $\mu=1.0$, $\eta=0.0$.

\textbf{Training Localization Expert}: On the CUB, we train our localization experts with a batch size of $bz=6$, an update rate $\lambda = 0.01$, and for $100$ epochs. The specific hyperparameters are as follows:
1) For VGG16: learning rate $lr=0.0005$, $\beta=0.5$, $\delta=0.0$.
2) For InceptionV3: $lr=0.0001$, $\beta=0.2$, $\delta=2e-8$.
3) For Resnet50: $lr=0.000125$, $\beta=0.5$, $\delta=0.0$.
On the ILSVRC, we train our localization experts with an update rate $\lambda = 0.001$ for $20$ epochs. The specific hyperparameters are as follows:
1) For VGG16: batch size $bz=84$, learning rate $lr=6e-5$, $\beta=1.0$, $\delta=1e-6$.
2)For InceptionV3: $bz=72$, $lr=0.0002$, $\beta=1.0$, $\delta=0.0$.
3) For Resnet50: $bz=24$, $lr=1.4e-5$, $\beta=1.0$, $\delta=0.0$.

\textbf{Training KG-CI-CAM}: On the CUB, we train our KG-CI-CAM with a batch size of $bz=6$, an update rate $\lambda = 0.01$, for $100$ epochs, and with a guiding hyper-parameter $\alpha=0.8$. The specific hyperparameters are as follows:
1) For VGG16: learning rate $lr=0.0005$, temperature $T^{cls}=T^{loc}=15$.
2) For InceptionV3: $lr=0.0001$, $T^{cls}=T^{loc}=15$.
3) For Resnet50: $lr=0.000125$, $T^{cls}=T^{loc}=10$.
On the ILSVRC, we train our KG-CI-CAM with a guiding hyper-parameter $\alpha=0.8$. The specific hyperparameters are as follows:
1) For VGG16: batch size $bz=78$, learning rate $lr=8e-5$, temperature $T^{cls}=T^{loc}=1.1$.
2) For InceptionV3: $bz=128$, $lr=9e-5$, $T^{cls}=T^{loc}=2$.
3) For Resnet50: $bz=32$, $lr=0.0001$, $T^{cls}=T^{loc}=5$.

During the testing phase, image resizing and central cropping are performed differently depending on the backbone used. For the VGG16 backbone, images are resized to $344 \times 344$ on the CUB ($288 \times 288$ on the ILSVRC) and subsequently centrally cropped to $224 \times 224$. This resizing and cropping strategy is inspired by~\cite{wu2022background, zhang2020rethinking}. When the backbone is InceptionV3, images undergo resizing to dimensions of $500 \times 500$ on the CUB ($404 \times 404$ on the ILSVRC) and are then centrally cropped to $299 \times 299$. Lastly, if the ResNet backbone is used, images are resized to $344 \times 344$ on the CUB ($304 \times 304$ on the ILSVRC) before undergoing central cropping to $224 \times 224$.

\subsection{Comparison with State-of-the-Art Methods}

\addtolength{\tabcolsep}{-0.3pt} 
\begin{table}[t]
   \centering
   \caption{Comparison with other state-of-the-art methods on the CUB-200-2011 dataset. $\ast$ indicates our baseline model and its re-implemented results by ourselves.}
   \label{tab:sota_acc_cub}
   \begin{tabular}{lcccc}
   \toprule
   Method & Backbone & Top-1 Cls & Top-1 Loc & GT-known\\
   \midrule
   NL-CCAM~\cite{yang2020combinational}$\ast$  &VGG16 &74.7 &65.6 &87.5    \\
   MEIL~\cite{mai2020erasing}   &VGG16        &  74.8       &    57.5  & 73.8   \\
   PSOL~\cite{zhang2020rethinking}&VGG16 & - & 66.3 & -  \\
   GCNet~\cite{lu2020geometry}&VGG16  & 76.8 & 63.2 & - \\
   RCAM~\cite{bae2020rethinking}&VGG16   &  74.9        &     61.3   & 80.7  \\
   MCIR~\cite{babar2021look}&VGG16 & 72.6 & 58.1 & - \\
   SLT-Net~\cite{guo2021strengthen}&VGG16  & 76.6 & 67.8& 87.6 \\
   SPA~\cite{pan2021unveiling}&VGG16 & - & 60.3 & 77.3 \\
   ORNet~\cite{xie2021online}&VGG16  & 77.0& 67.7 & - \\
   BridgeGap~\cite{kim2022bridging}&VGG16 & - & \blue{70.8} & \textbf{93.2}   \\
   CREAM~\cite{xu2022cream}&VGG16 & - & 70.4 & 91.0 \\
   \rowcolor{mygray}KG-CI-CAM&VGG16 & \textbf{79.2} &	\textbf{73.0} &	\blue{91.6} \\ 
   \hline
 NL-CCAM~\cite{yang2020combinational}$\ast$  &Resnet50   &82.8 &78.6 &94.7      \\   PSOL~\cite{zhang2020rethinking}&Resnet50 &-&70.7&\blue{90.0}\\
   RCAM~\cite{bae2020rethinking}&Resnet50 &75.0&59.5&77.6\\
   CREAM~\cite{xu2022cream}&Resnet50 &-&\blue{76.0}&89.9\\
   \rowcolor{mygray}KG-CI-CAM&Resnet50 &\textbf{84.0}&\textbf{81.6}&\textbf{96.8}\\	
   \hline
    NL-CCAM~\cite{yang2020combinational}$\ast$  &InceptionV3   &77.6 &71.4 &91.8      \\ PSOL~\cite{zhang2020rethinking}&InceptionV3 &-&65.5&-\\
   GC-Net~\cite{lu2020geometry}&InceptionV3 &-&58.6&75.3\\	
   I$^2$C~\cite{zhang2020inter}&InceptionV3 &-&55&72.6\\
   SLT-Net~\cite{guo2021strengthen}&InceptionV3 &76.4&66.1&86.5\\
   SPA~\cite{pan2021unveiling}&InceptionV3 &-&53.6&72.1\\
   CREAM~\cite{xu2022cream}&InceptionV3 &-&\blue{71.8}&\blue{90.4}\\
   \rowcolor{mygray}KG-CI-CAM&InceptionV3 &\textbf{79.8}& \textbf{76.3}& \textbf{95.3} \\
   \hline

   \hline
   \end{tabular}
\end{table}
\addtolength{\tabcolsep}{0.3pt} 

\addtolength{\tabcolsep}{2.7pt} 
\begin{table}[t]
   \centering
   \caption{Quantitative results using threshold-independent MaxBoxAccV2 metric. $+$ ($-$) numbers denote the absolute increase (decrease) over CAM.}
   \label{tab:sota_MaxBoxAccV2}
   \begin{tabular}{lcccc}
      \toprule
      \multirow{2}{*}{Method}& \multicolumn{4}{c}{CUB-200-2011 (MaxBoxAccV2)} \\
      & VGG16 & InceptionV3 & Resnet50 & Mean \\
      
      \hline
      CAM~\cite{zhou2016learning} & 63.7 & 56.7 & 63.0 & 61.1\\

      HaS~\cite{kumar2017hide}& \lightgreen{+0.0} & \red{-3.3} & \lightgreen{+1.7} & \red{-0.5}  \\
      
      ACoL~\cite{zhang2018adversarial}& \red{-6.3} & \red{-0.5} & \lightgreen{+3.5} & \red{-1.1} \\
      
      SPG~\cite{zhang2018self} & \red{-7.4} & \red{-0.8} & \red{-2.6} & \red{-3.6} \\
      
      ADL~\cite{choe2019attention} & \lightgreen{+2.6} & \lightgreen{+2.1} & \red{-4.6} & \lightgreen{+0.0} \\
      
      CutMix\cite{yun2019cutmix} & \red{-1.4} & \lightgreen{+0.8} & \red{-0.2} & \red{-0.3} \\
      
      CAM\_IVR\cite{kim2021normalization} &  \lightgreen{+1.5} &  \lightgreen{+4.1} & \lightgreen{+3.9} & \lightgreen{+3.1} \\

      CREAM~\cite{xu2022cream} &  \textbf{\lightgreen{+7.8}} & \lightgreen{+7.5} & \lightgreen{+10.5} & \lightgreen{+8.6} \\
      \hline
      KG-CI-CAM & \lightgreen{+5.9} & \textbf{\lightgreen{+11.1}} & \textbf{\lightgreen{+15.0}} & \textbf{\lightgreen{+10.7}} \\
      \bottomrule
   \end{tabular}
\end{table}
\addtolength{\tabcolsep}{-2.7pt}

\addtolength{\tabcolsep}{-0.7pt} 
\begin{table}[t]
   \centering
   \caption{Comparison with other state-of-the-art methods on the ILSVRC 2016 dataset. $\ast$ indicates our baseline model and its re-implemented results by ourselves.}
   \label{tab:sota_acc_imagenet}
   \begin{tabular}{lcccc}
   \toprule
   Method & Backbone & Top-1 Cls & Top-1 Loc & GT-known\\
   \midrule
   NL-CCAM~\cite{yang2020combinational}$\ast$  &VGG16      &   72.3      &    48.6 & 62.9   \\
   MEIL~\cite{mai2020erasing}   &VGG16          &   70.3        &     46.8   & -   \\
   PSOL~\cite{zhang2020rethinking}&VGG16  &  - & 50.9 & 64.0 \\
   RCAM~\cite{bae2020rethinking}&VGG16     &   67.2        &      45.4   & 62.7\\
   MCIR~\cite{babar2021look}&VGG16  &  71.2 & 51.6 & 66.3\\
   SLT-Net~\cite{guo2021strengthen}&VGG16  &  \textbf{72.4} & 51.2 & 67.2\\
   SPA~\cite{pan2021unveiling}&VGG16  &  - & 49.6 & 65.1\\
   ORNet~\cite{xie2021online}&VGG16   &  71.6 & \blue{52.1} & -\\
   BridgeGap~\cite{kim2022bridging}&VGG16   &  - & 49.9 & \textbf{68.9} \\
   CREAM~\cite{xu2022cream}&VGG16  & - & \textbf{52.4} & 68.3\\
   \rowcolor{mygray}KG-CI-CAM&VGG16  & \textbf{72.4}	& 51.4 & 67.4 \\ 
   \hline
    NL-CCAM~\cite{yang2020combinational}$\ast$  &Resnet50      &   74.2      &    53.2 & 68.3   \\
   PSOL~\cite{zhang2020rethinking}&Resnet50 &-&54&65.4\\
   RCAM~\cite{bae2020rethinking}&Resnet50 &\textbf{75.8}&49.4&62.2\\
   CREAM~\cite{xu2022cream}&Resnet50 &-&\textbf{55.7}&\blue{69.3}\\
\rowcolor{mygray}KG-CI-CAM&Resnet50 & 74.2 & 54.5 &\textbf{70.1}\\
   \hline
   NL-CCAM~\cite{yang2020combinational}$\ast$  &InceptionV3      &   73.1      &    52.9 & 66.8   \\
   PSOL~\cite{zhang2020rethinking}&InceptionV3 &-&54.8&65.2\\
   MEIL~\cite{mai2020erasing}&InceptionV3 &73.3&49.5&-\\
   GC-Net~\cite{lu2020geometry}&InceptionV3 &77.4&49.1&-\\	
   I$^2$C~\cite{zhang2020inter}&InceptionV3&73.3&53.1&68.5\\
   SLT-Net~\cite{guo2021strengthen}&InceptionV3 &\textbf{78.1}&\blue{55.7}&67.6\\
   SPA~\cite{pan2021unveiling}&InceptionV3 &-&52.7&68.3\\
   CREAM~\cite{xu2022cream}&InceptionV3 &-&\textbf{56.1}&\blue{69}\\
   \rowcolor{mygray}KG-CI-CAM&InceptionV3 & 74.5 & 55.0 &\textbf{70.3}\\
   \hline

   \hline
   \end{tabular}
\end{table}
\addtolength{\tabcolsep}{0.7pt} 

We conduct a comparison of KG-CI-CAM with other state-of-the-art (SOTA) methods on both the CUB-200-2011~\cite{wah2011caltech} and ILSVRC 2016~\cite{russakovsky2015imagenet} datasets.

On the CUB-200-2011 dataset, our evaluation revealed that KG-CI-CAM outperforms the current SOTA method across various evaluation metrics, as demonstrated in Table~\ref{tab:sota_acc_cub}. Specifically, when utilizing VGG16~\cite{simonyan2014very} as the backbone, KG-CI-CAM respectively increases the Top-1 Cls, Top-1 Loc, and GT-Known by $4.5\%$, $7.4\%$, and $4.1\%$ over the baseline. Besides, KG-CI-CAM achieves a Top-1 classification accuracy of $79.2\%$, which is $2.2\%$ higher than the current SOTA method ORNet~\cite{xie2021online}. Moreover, it surpasses ORNet~\cite{xie2021online} by $5.3\%$ in terms of Top-1 localization accuracy. In comparison to the GT-known localization SOTA method BridgeGap~\cite{kim2022bridging}, KG-CI-CAM demonstrates a slight decrease of $1.6\%$ in the GT-known localization accuracy but delivers a substantial gain of $2.2\%$ in the Top-1 localization accuracy. When employing InceptionV3~\cite{szegedy2016rethinking} or Resnet50~\cite{he2016deep} as the backbone, KG-CI-CAM consistently brings significant improvements over baseline and outperforms other methods. For instance, with InceptionV3, KG-CI-CAM achieves an additional $3.4\%$ in the Top-1 classification accuracy, $4.5\%$ in the Top-1 localization accuracy, and $4.9\%$ in the GT-known localization accuracy compared to SOTA methods. Similarly, when using Resnet50 as the backbone, KG-CI-CAM exhibits a significant improvement, achieving an extra $9.0\%$ in the Top-1 classification accuracy, $5.6\%$ in the Top-1 localization accuracy, and $6.8\%$ in the GT-known localization accuracy compared to SOTA methods. Besides, our results in Table~\ref{tab:sota_MaxBoxAccV2} demonstrate that KG-CI-CAM also excels in the MaxBoxAccV2~\cite{choe2020evaluating}, surpassing the current SOTA method CREAM~\cite{xu2022cream} by $2.1\%$ on average.

In more general scenarios, such as the ILSVRC 2016~\cite{russakovsky2015imagenet} dataset, where the issue of ``entangled context'' is less prevalent due to the large number of images and diverse backgrounds, the performance of KG-CI-CAM is not as pronounced as on the CUB-200-2011 dataset. Nevertheless, it still achieves SOTA performance on certain metrics, which is shown in Table~\ref{tab:sota_acc_imagenet}. For instance, when using VGG16~\cite{simonyan2014very} as the backbone, KG-CI-CAM increases the Top-1 Loc and GT-Known by $2.8\%$ and $4.5\%$ over the baseline. Besides, KG-CI-CAM achieves a SOTA classification accuracy, matching the performance of SLT-Net~\cite{guo2021strengthen}. Compared to SLT-Net~\cite{guo2021strengthen}, KG-CI-CAM respectively achieves a $0.2\%$ improvement in the Top-1 localization accuracy and GT-known localization accuracy. When the backbone is InceptionV3~\cite{szegedy2016rethinking} or Resnet50~\cite{he2016deep}, KG-CI-CAM achieves a new SOTA in the GT-known localization. Although KG-CI-CAM can not achieve SOTA performance in the Top-1 Loc, it brings significant improvements over the baseline. In addition, it is important to note that our baseline model's lower Top-1 Cls score affects the Top-1 Loc performance. Since Top-1 Loc is influenced by both Top-1 Cls and GT-known, the phenomenon of high GT-known and low Top-1 Loc on the ILSVRC 2016 dataset is primarily due to the characteristics of our baseline. 



\subsection{Ablation Study}
\addtolength{\tabcolsep}{-3pt} 
\begin{table}[t]
   \centering
   \caption{Ablation studies on the CUB-200-2011 dataset. Baseline performance is our re-implemented results. 1) Baseline: NL-CCAM, 2) Causal: Causal Intervention. 3) Knowledge: Knowledge Guidance. 4) AccV2: MaxBoxAccV2.}
   \label{tab:ablation_components}
   \begin{tabular}{ccc|cccc}
   \hline

   \hline                                             
   Baseline & Causal & Knowledge & Top-1 Cls & Top-1 Loc & GT-known   & AccV2   \\ 
   
   \hline
   \multicolumn{7}{c}{Backbone: VGG16} \\  \hline  
   $\surd$ &         &         & 74.7 & 65.6 & 87.5 & 65.7  \\
   $\surd$ & $\surd$ &         & 75.1 & 66.6 & 88.1 & 66.9  \\
   $\surd$ & $\surd$ & $\surd$ & \textbf{79.2} & \textbf{73.0} & \textbf{91.6} & \textbf{69.6} \\
   \hline
   \multicolumn{7}{c}{Backbone: InceptionV3} \\  \hline  
   $\surd$ &         &         & 77.6 & 71.4 & 91.8 & 64.1  \\
   $\surd$ & $\surd$ &         & 78.7 & 73.9 & 93.8 & 65.5  \\
   $\surd$ & $\surd$ & $\surd$ & \textbf{79.8} & \textbf{76.3} & \textbf{95.3} & \textbf{67.8}  \\
   \hline
   \multicolumn{7}{c}{Backbone: Resnet50} \\  \hline  
   $\surd$ &         &         & 82.8 & 78.6 & 94.7 & 74.6 \\
   $\surd$ & $\surd$ &         & 82.9 & 79.1 & 95.2 & 76.9  \\
   $\surd$ & $\surd$ & $\surd$ & \textbf{84.0} & \textbf{81.6} & \textbf{96.8} & \textbf{78.0}  \\
   \hline

   \hline
   \end{tabular}
\end{table}
\addtolength{\tabcolsep}{3pt} 

In this section, we have conducted three sets of ablation studies on the CUB-200-2011~\cite{wah2011caltech} dataset, employing three different backbones to both quantitatively and qualitatively showcase the efficacy of our proposed methods.

\begin{figure*}[t]
   \centering
   \includegraphics[width=0.95\linewidth]{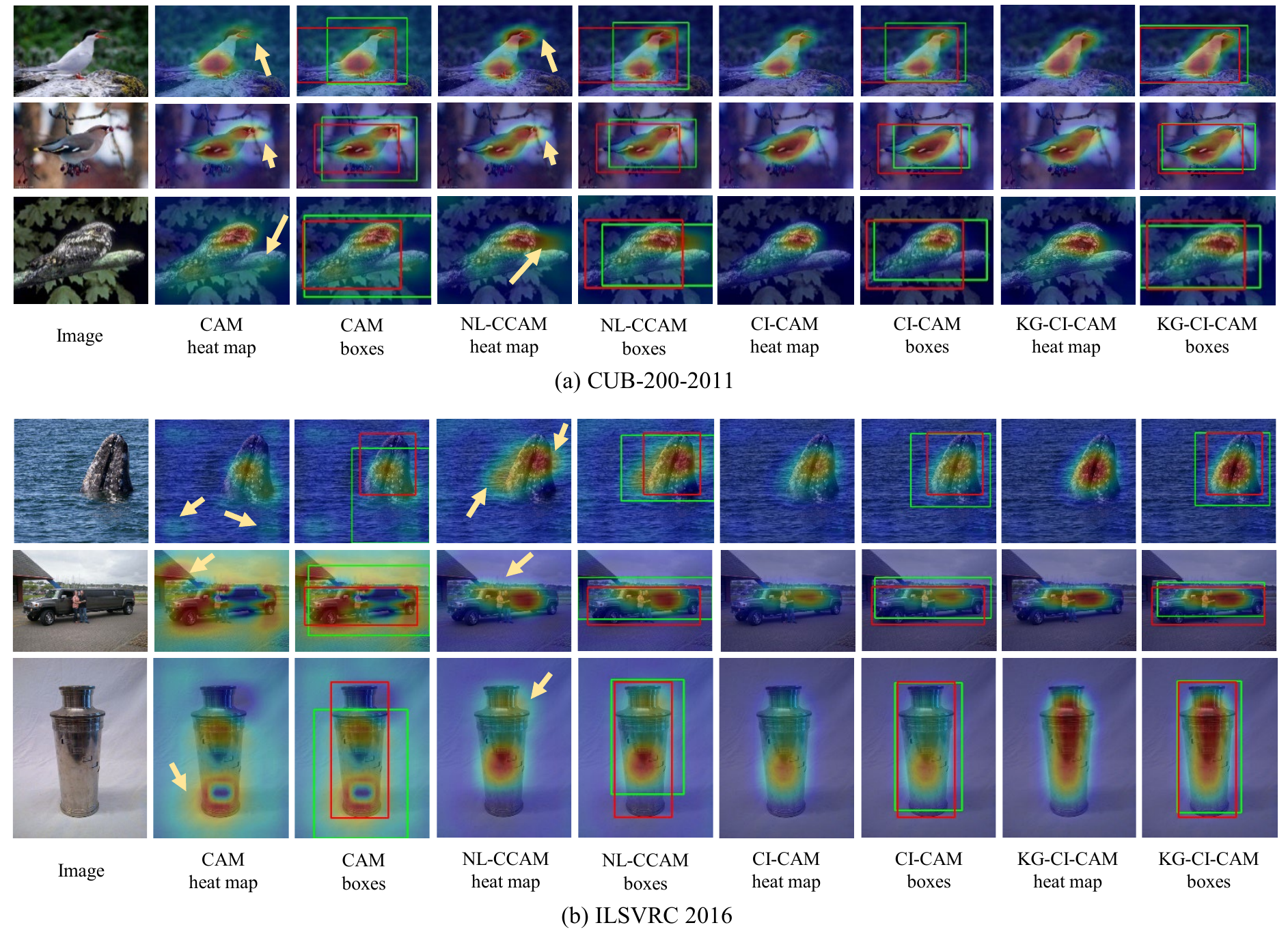}
   \caption{Qualitative object localization results compared with the CAM and NL-CCAM methods. The predicted bounding boxes are in green, and the ground-truth boxes are in red. The yellow arrows indicate the regions suffer from entangled contexts.}
   \label{result_images}
\end{figure*}

\textbf{Causal Intervention for ``Entangled Context'' Issue:} In terms of quantitative analysis, training the baseline model with causal intervention yields comprehensive enhancements across all evaluation metrics listed in Table~\ref{tab:ablation_components}. Notably, it results in improved localization and classification performance. For instance, with VGG16~\cite{simonyan2014very} as the backbone, causal intervention brings about enhancements of $0.4\%$, $1.0\%$, $0.6\%$, and $1.2\%$ in the Top-1 Cls, Top-1 Loc, GT-known Loc, and MaxBoxAccV2 metrics, respectively. Similarly, when the backbone is InceptionV3~\cite{szegedy2016rethinking}, causal intervention leads to improvements of $1.1\%$, $2.5\%$, $2.0\%$, and $1.4\%$ in the Top-1 Cls, Top-1 Loc, GT-known Loc, and MaxBoxAccV2 metrics, respectively. Furthermore, with Resnet50~\cite{he2016deep} as the backbone, causal intervention leads to improvements of $0.1\%$, $0.5\%$, $0.5\%$, and $2.3\%$ in the respective metrics. Qualitative analysis in Figure~\ref{result_images} illustrates that our CI-CAM method, augmented by causal intervention, displays the clearer distinction between object and co-occurring background compared to vanilla CAM~\cite{zhou2016learning} and NL-CCAM~\cite{yang2020combinational}, thereby vividly demonstrating the effectiveness of causal intervention in addressing the ``entangled context'' issue.

\addtolength{\tabcolsep}{-2.5pt} 
\begin{table}[t]
   \centering
   \caption{Analysis experiments of multi-source knowledge on the CUB-200-2011 dataset. 1) ClsExp: Classification Expertise. 2) LocExp: Localization Expertise. 3) AccV2: MaxBoxAccV2.}
   \label{tab:multi-source knowledge}
   \begin{tabular}{ccc|cccc}
   \hline

   \hline                                             
   CI-CAM & ClsExp & LocExp & Top-1 Cls & Top-1 Loc & GT-known   &   AccV2   \\ 
   
   \hline
   \multicolumn{7}{c}{Backbone: VGG16} \\  \hline  
   $\surd$ &         &         & 75.1 & 66.6 & 88.1 & 66.9  \\
   $\surd$ & $\surd$ &         & 77.4 & 66.5 & 85.0 & 65.2  \\
   $\surd$ &         & $\surd$ & 78.1 & 71.8 & 91.5 & \textbf{69.6} \\
   $\surd$ & $\surd$ & $\surd$ & \textbf{79.2} & \textbf{73.0} & \textbf{91.6} & \textbf{69.6}  \\
   \hline
   \multicolumn{7}{c}{Backbone: InceptionV3} \\  \hline  
   $\surd$ &         &         & 78.7 & 73.9 & 93.8 & 65.5   \\
   $\surd$ & $\surd$ &         & \textbf{79.8} & 74.3 & 92.7 & 64.8  \\
   $\surd$ &         & $\surd$ & 78.6 & 74.9 & 94.9 & 67.3  \\
   $\surd$ & $\surd$ & $\surd$ & \textbf{79.8} & \textbf{76.3} & \textbf{95.3} & \textbf{67.8} \\
   \hline
   \multicolumn{7}{c}{Backbone: Resnet50} \\  \hline  
   $\surd$ &         &         & 82.9 & 79.1 & 95.2 & 76.9  \\
   $\surd$ & $\surd$ &         & \textbf{84.0} & 80.8 & 95.7 & 75.0 \\
   $\surd$ &         & $\surd$ & 83.1 & 80.5 & 96.6 & 76.8  \\
   $\surd$ & $\surd$ & $\surd$ & \textbf{84.0} & \textbf{81.6} & \textbf{96.8} & \textbf{78.0} \\
   \hline

   \hline
   \end{tabular}
\end{table}
\addtolength{\tabcolsep}{2.5pt} 

\begin{table*}[t]
   \centering
   \caption{Analysis experiments of using different localization experts on the CUB-200-2011 dataset.}
   \label{tab:loc_experts}
   \begin{tabular}{l|cccc|ccc}
   \hline
      
   \hline
   \multicolumn{1}{l|}{\multirow{2}{*}{Backbone}} & \multicolumn{4}{c|}{Localization Expert} & \multicolumn{3}{c}{KG-CI-CAM}                                                                  \\
   
   \multicolumn{1}{c|}{} & \multicolumn{1}{l}{Expert Alias}  & Top-1 Cls & Top-1 Loc & GT-known &  Top-1 Cls & Top-1 Loc & GT-known \\ 
   
   \hline
   \multirow{2}*{\tabincell{l}{VGG16}} & \multicolumn{1}{l}{Best Top-1 Loc Expert}  & \textbf{77.0} & \textbf{68.3} & 87.9 & 78.3 & 71.5 & 90.5 \\
   & \multicolumn{1}{l}{Best GT-known Loc Expert} & 73.3 & 65.9 & \textbf{88.8} & \textbf{79.2} & \textbf{73.0} & \textbf{91.6} \\
   \hline
   \multirow{2}*{\tabincell{l}{InceptionV3}} & \multicolumn{1}{l}{Best Top-1 Loc Expert}  & \textbf{73.5} & \textbf{68.6} & 92.9 & 79.1 & 75.0 & 94.5 \\
   & \multicolumn{1}{l}{Best GT-known Loc Expert} & 72.1 & 68.2 & \textbf{94.2} & \textbf{79.7} & \textbf{76.3} & \textbf{95.3} \\
   \hline
   \multirow{2}*{\tabincell{l}{Resnet50}} & \multicolumn{1}{l}{Best Top-1 Loc Expert}  & \textbf{80.2} & \textbf{76.7} & 95.2 & 83.9 & 81.2 & 96.5 \\
   & \multicolumn{1}{l}{Best GT-known Loc Expert} & 78.9 & 75.8 & \textbf{95.7} & \textbf{84.0} & \textbf{81.6} & \textbf{96.8} \\
   \hline

   \hline
   \end{tabular}
\end{table*}

\textbf{Multi-source Knowledge Guidance for ``C-L Dilemma" Issue:} In terms of quantitative analysis, training the causal model with multi-source knowledge guidance also results in notable improvements across all evaluation metrics in Table~\ref{tab:ablation_components}. These enhancements encompass not only localization performance but also classification performance. For instance, employing VGG16~\cite{simonyan2014very} as the backbone, knowledge guidance results in enhancements of $4.1\%$, $6.4\%$, $3.5\%$, and $2.7\%$ in the Top-1 Cls, Top-1 Loc, GT-known Loc, and MaxBoxAccV2 metrics, respectively. Likewise, with InceptionV3~\cite{szegedy2016rethinking} as the backbone, knowledge guidance brings about improvements of $1.1\%$, $2.4\%$, $1.5\%$, and $2.3\%$ in the respective metrics. Additionally, with Resnet50~\cite{he2016deep} as the backbone, knowledge guidance leads to enhancements of $1.1\%$, $2.5\%$, $1.6\%$, and $1.1\%$ in the respective metrics. Qualitative analysis in Figure~\ref{result_images} demonstrates that our KG-CI-CAM method, equipped with knowledge guidance, more effectively distinguishes foreground and background, emphasizing foreground regions. This vividly underscores the positive impact of knowledge guidance in the context of the WSOL task.

\subsection{Analysis and Discussion}
In this section, we undertake analysis and discussion on the required knowledge guidance and the selection of appropriate knowledge sources to guide our model's training on the CUB-200-2011~\cite{wah2011caltech} dataset. Initially, we perform three sets of experiments through using three different backbones to elucidate the types of knowledge that prove crucial. Subsequently, considering that in the evaluation metric of Accuracy, Top-1 Loc is influenced by classification performance while GT-known remains unaffected, we conduct an analytical experiment to ascertain the optimal selection of localization experts.

\textbf{Required Knowledge Guidance:} To unveil the indispensable knowledge, we set up a series of experiments utilizing various knowledge on the CUB-200-2011~\cite{wah2011caltech} dataset, the outcomes of which are reported in Table~\ref{tab:multi-source knowledge}. Our observations demonstrate that incorporating classification expertise yields significant enhancement in classification performance across all backbone configurations. Moreover, integrating localization expertise leads to comprehensive improvements in localization performance. Notably, simultaneous utilization of both classification and localization expertise yields even more substantial improvements in both classification and localization. Taking VGG16~\cite{simonyan2014very} as the backbone, employing only the classification expertise results in a $2.3\%$ improvement in the Top-1 Cls. Utilizing solely the localization expertise leads to $5.3\%$, $3.4\%$, and $2.7\%$ enhancements in the Top-1 Loc, GT-known Loc, and MaxBoxAccV2, respectively. When integrating both expertise, further improvements are recorded, achieving $4.1\%$, $6.4\%$, $3.5\%$, and $2.7\%$ enhancements in the Top-1 Cls, Top-1 Loc, GT-known Loc, and MaxBoxAccV2. Similar trends are observed when InceptionV3~\cite{szegedy2016rethinking} or Resnet50~\cite{he2016deep} are used as the backbones. These results underscore the advantage of leveraging a combination of classification and localization expertise to enhance performance.

\textbf{Optimal Localization Expert Selection:} Taking the example of the CUB-200-2011~\cite{wah2011caltech} dataset, we train KG-CI-CAM twice using the same classification expert and two distinct localization experts, as detailed in Table~\ref{tab:loc_experts}. When VGG16~\cite{simonyan2014very} is the backbone, we find that the localization knowledge provided by the Best GT-known Loc Expert proves more effective than that of the Best Top-1 Loc Expert. This observation remains consistent even when different backbones, such as InceptionV3~\cite{szegedy2016rethinking} and Resnet50~\cite{he2016deep}, are employed. This consistency across models substantiates the recommendation that the choice of a localization expert should be primarily based on its performance in the GT-known Loc, rather than the Top-1 Loc performance.

\section{Conclusions}
In this paper, our focus centers on the unaddressed and overlooked challenges of the ``entangled context'' and ``C-L dilemma'' within the Weakly-supervised Object Localization (WSOL) task. To tackle these issues comprehensively, we introduce a systematic framework that simultaneously addresses both problems. Specifically, we initiate our approach by addressing the ``entangled context'' problem through causal intervention. This involves analyzing the causal dynamics between image features, context, and image labels, thereby neutralizing the impact of confounding context on image features as shown in Figure~\ref{structural_causal_model_graph}. In terms of the model architecture, we introduce a causal context pool that accumulates context information for each class, which is then projected onto convolutional layer feature maps to enhance feature clarity, which is shown in Figure~\ref{network_architecture_graph}. Subsequently, we devise a multi-source knowledge guidance framework, illustrated in Figure~\ref{fig:kd} (a), to mitigate the ``C-L dilemma''. This framework facilitates the assimilation of classification and localization knowledge during model training. Notably, our study marks one of the initial attempts at comprehending and addressing the intertwined challenges of the ``entangled context'' and ``C-L dilemma'' in the realm of WSOL. Thorough experimentation validates the practical existence of these challenges within the WSOL task and underscores the efficacy of our proposed solutions, as evidenced in Figure~\ref{comparison_graph} and Figure~\ref{result_images}.

\section*{Acknowledgments}

This work was supported by the National Natural Science Foundation of China (62337001, 62293554, 62206249, U2336212), Natural Science Foundation of Zhejiang Province, China (LZ24F020002), Young Elite Scientists Sponsorship Program by CAST (2023QNRC001), and the Fundamental Research Funds for the Central Universities(No. 226-2022-00051).




{\footnotesize
\bibliographystyle{plain}
\bibliography{main}
}

\vfill

\end{document}